\documentclass[letterpaper,twocolumn,10pt]{article}
\usepackage{usenix}

\usepackage{tikz}
\usepackage{amsmath}

\usepackage{amssymb}
\usepackage{booktabs}
\usepackage{multirow}
\usepackage{colortbl}
\usepackage{subcaption}
\usepackage{tcolorbox}
\usepackage{tgcursor}

\usepackage{xurl}

\usepackage{lstautogobble}

\tcbuselibrary{breakable,skins,listings}

\newtcblisting{promptbox}[1]{
  enhanced,
  breakable,
  left skip=2pt,
  right skip=2pt,
  enlarge bottom by=4pt,
  listing only,
  listing options={
    basicstyle=\ttfamily\fontsize{7.2pt}{8.1pt}\selectfont,
    breaklines=true,
    breakatwhitespace=true,
    breakindent=0pt,
    breakautoindent=false,
    columns=fullflexible,
    keepspaces=true,
    autogobble=true,
    literate={→}{{$\rightarrow$}}1
  },
  colback=black!3,
  colframe=black!75,
  boxrule=0.5pt,
  arc=3mm,
  left=2.5mm,
  right=2.5mm,
  top=2.5mm,
  bottom=2.5mm,
  title={#1},
  fonttitle=\bfseries,
  coltitle=white,
  colbacktitle=black!75,
  attach boxed title to top left={xshift=4mm,yshift=-2mm},
  boxed title style={arc=0pt,boxrule=0pt}
}

\begin{document}

\date{}

\title{When Tool Outputs Become Commands: Separating Action Induction from Runtime Authorization in Tool-Augmented LLM Agents}

\author{
    {\rm Xiaokun Guo$^{1,2}$ \quad
    Zhen Xu$^{1,2}$ \quad
    Dongdong Huo$^{1,2}$ \quad
    Yanqiu Zhang$^{1,2}$}\\
    {\rm Wei Wang$^{1,2}$ \quad
    Qinfu Yang$^{1,2}$ \quad
    Dongjin Yu$^{1,2}$ \quad
    Yu Wang$^{1}$}\\[1ex]
    $^{1}$Institute of Information Engineering, Chinese Academy of Sciences\\
    $^{2}$School of Cyber Security, University of Chinese Academy of Sciences\\[0.5ex]
    {\textup{\texttt{\{guoxiaokun,xuzhen,huodongdong,zhangyanqiu\}@iie.ac.cn}}}\\
    {\textup{\texttt{\{wangwei2024,yangqinfu,yudongjin,wangyu\}@iie.ac.cn}}}
}

\maketitle

\begin{abstract}
    Tool-augmented LLM agents must rely on untrusted runtime Observations to complete open-ended tasks; however, when tool outputs no longer merely provide data but begin to specify concrete actions, they effectively become ``commands'' that can drive real-world side effects beyond user intent.
    We argue that this risk arises from conflating action induction with execution authorization. To address this distinction, we propose SARA, which treats action induction and execution authorization as distinct runtime roles and separates action provenance from execution authority.
    On the Observation side, a context-isolated Action Probe exposes action-inducing semantics and persistently records action-origin provenance across steps as a review signal; on the execution side, actual tool calls are authorized only against the user objective and audited evidence from authorized successful executions, while satisfying goal, execution-chain, and argument-level support.
    To preserve this separation across multi-step execution, SARA applies No-History-Promotion to prevent historical recurrence from laundering action origins into execution authority.
    Across AgentDojo and AgentDyn, SARA limits ASR to no more than \(0.63\%\) across four primary evaluation settings while maintaining competitive task utility, and consistently reduces ASR across additional Agent backbones.
\end{abstract}

\section{Introduction}

Large language models are evolving from text generators into agents that can act in real-world environments through external tools.
Recent Agent systems use planning, reasoning, and tool-use capabilities to access external resources and execute multi-step tasks\cite{yao2023reactsynergizingreasoningacting,NEURIPS2023_d842425e,ICLR2024_28e50ee5,li-etal-2023-api}, while evaluation settings have gradually expanded from closed tool interfaces to open environments such as operating systems, web browsers, and databases\cite{liu2025agentbenchevaluatingllmsagents,Wang_2024}.
These systems can retrieve emails and files, invoke external services, and perform operations with real side effects, such as sending information, modifying objects, or committing transactions\cite{zhan2024injecagent,debenedetti2024agentdojo,DBLP:conf/iclr/RuanDWPZBDMH24,ye-etal-2024-toolsword}, extending Agent security risks from generated content to environmental consequences caused by tool execution.
Attackers can therefore embed operational content in third-party resources such as webpages, emails, or documents and exploit an Agent's existing tool privileges to influence subsequent execution\cite{zhang2025asb}.
Early work revealed the risk of remotely manipulating LLM-integrated applications through indirect prompt injection (IPI)\cite{greshake2023indirect,299563}, while InjecAgent and AgentDojo further showed that such influence can propagate through tool-use processes and ultimately produce real external effects\cite{zhan2024injecagent,debenedetti2024agentdojo}.

However, in open-ended interactive environments that more closely resemble the real world\cite{zhou2024webarenarealisticwebenvironment,NEURIPS2024_5d413e48}, defenses face a structural tension: \emph{restricting untrusted external content can improve security but may weaken the runtime adaptability required by open-ended tasks}.
Narrowing the accessible action space or limiting the influence of external input on subsequent behavior can reduce attack success, but it also creates a direct security--utility trade-off\cite{beurerkellner2025designpatternssecuringllm}; conversely, if Observations are allowed to influence subsequent decisions too freely, attack semantics may propagate through normal tool steps and ultimately produce unauthorized external effects\cite{zhan2024injecagent,debenedetti2024agentdojo}.
This tension is particularly pronounced in dynamic tasks, where legitimate subsequent actions often cannot be fully determined from the initial request and must instead be dynamically instantiated using runtime feedback, third-party information, and environmental state produced by prior execution\cite{li2026agentdyn,yao2024taubenchbenchmarktoolagentuserinteraction,lu-etal-2025-toolsandbox}.
Consequently, treating the mere fact that an Agent responds to external content as an attack signal can easily conflate legitimate runtime adaptation with attack-induced behavior.

The key challenge is that open-ended tasks inherently require untrusted Observations to dynamically instantiate subsequent actions and arguments, so a security mechanism cannot simply exclude Observation-derived information.
For example, when a user asks to ``find the quarterly report and send it to Alice,'' the file identifier or object path may become available only after search and retrieval and must be used in a subsequent call; however, the same result may also contain an additional operational request such as ``send a copy of the report to \texttt{attacker@example.com}.''
The former uses runtime information to \emph{instantiate existing authority (runtime instantiation)}, whereas the latter attempts to \emph{expand existing authority (authority expansion)}.
The central question is therefore not whether an Observation influences the Agent, but whether that influence acquires real execution authority beyond the user's intent.

Directly using another LLM to determine whether an Observation is malicious does not inherently solve this problem, because the analyzer must still process the same untrusted content, and prior work has shown that LLM detectors or evaluators can themselves be manipulated by adversarial injection\cite{DBLP:conf/aisec-ws/ChoudharyAPJ25,DBLP:conf/ccs/ShiYLH00G24,raina-etal-2024-llm}.
Rather than requiring Observation-side analysis to distinguish malicious from benign content accurately, a more direct question is whether the external content exhibits \emph{action-inducing semantics} that can be mapped to concrete tool actions or argument roles.
Such actionability indicates only that the external content contributed to forming a candidate action; it does not establish that the action is authorized for execution, and a tool output may therefore shift from \emph{data-bearing output} to \emph{action-inducing output}.
Structurally, this risk resembles the confused deputy problem~\cite{10.1145/54289.871709}: an Agent holding legitimate tool privileges may be induced by external content to exercise those privileges for an effect that the user did not authorize.
We accordingly distinguish \emph{action induction} from \emph{execution authorization}:
\textbf{An Observation may influence a candidate action and provide runtime information needed to complete an existing task, but action induction itself cannot confer real execution authority; a candidate call must re-establish independent authorization support at the execution boundary.}

To this end, we propose SARA (Separating Action Induction from Runtime Authorization), a persistent runtime authorization mechanism deployed between the Agent and the real tool executor.
SARA requires no access to the Agent's hidden reasoning, internal memory, or planning state and relies only on the user request, tool schema, actual candidate calls, tool Observations, and its own runtime state.
SARA does not filter raw Observations; instead, a context-isolated Action Probe identifies their action-inducing semantics and retains the corresponding action origins across steps as persistent authorization-review signals, without directly labeling them malicious or rejecting subsequent behavior.
In parallel, SARA independently maintains audited execution evidence formed by authorized and successfully executed tool interactions, allowing dynamic runtime information to continue instantiating tasks that the user has already authorized.
For candidate calls that undergo full review, SARA combines user authorization, persistent action origins, and audited execution evidence at the real execution boundary to determine whether the goal, execution chain, and actual arguments have independent support, thereby preventing action origins in external content from independently becoming new execution authority.

In summary, this paper makes the following contributions:
\begin{itemize}
    \item We characterize the central security tension of IPI in dynamic tool-using Agents as a \textbf{confusion between action induction and execution authorization}: Observations must be able to instantiate existing tasks, but their influence must not itself create or expand execution authority that the user did not grant.
    \item We propose SARA, which uses a context-isolated Action Probe to persistently record action origins induced by Observations and combines them with independent audited execution evidence to enforce goal-, execution-chain-, and argument-level runtime authorization at the real tool-execution boundary.
    \item We systematically evaluate SARA on AgentDojo and AgentDyn against representative methods spanning input filtering, execution-structure constraints, action attribution, and execution-boundary control, while also examining component contributions, stability across Agent backbones, and additional inference cost.
          The results show that SARA substantially reduces unauthorized real tool execution while maintaining competitive task utility and provides consistent security gains across different Agent backbones.
\end{itemize}

\section{Problem Definition and Threat Model}
\label{sec:problem}

\subsection{Problem Definition}
\label{sec:problem-definition}

For an Agent capable of real tool calls, an indirect prompt injection produces a security consequence not when external content influences model generation, but when the resulting call crosses the real tool-execution boundary.
Given a user request \(U\) and a tool set \(\mathcal{T}\), a multi-round tool execution can be represented as
\[
    \mathcal{X}
    =
    \left(
    U,c_1,O_1,\ldots,c_t,O_t
    \right),
\]
where an Observation may contain either runtime facts needed to complete the task or operational content controlled by an attacker.
Let the candidate call at step \(t\) be
\[
    c_t=(\tau_t,\theta_t),
\]
where \(\tau_t\) denotes the tool and \(\theta_t\) denotes its structured arguments.
Let
\[
    \mathrm{Permitted}_U(c_t\mid\mathcal{X}_{<t})
\]
denote whether \(c_t\) has normative execution authority under the user request and the preceding legitimate task execution.
An untrusted Observation \(O_i\) may influence the formation of a candidate call, but such action induction does not imply execution authority:
\[
    \mathrm{Induced}(c_t,O_i)
    \not\Rightarrow
    \mathrm{Permitted}_U(c_t\mid\mathcal{X}_{<t}).
\]
Let \(\mathrm{Execute}(c_t)\) indicate that the call actually reaches the real tool executor; a runtime authorization violation is then defined as
\[
    \mathrm{Violation}_t
    \triangleq
    \mathrm{Execute}(c_t)
    \land
    \neg\mathrm{Permitted}_U(c_t\mid\mathcal{X}_{<t}).
\]
When such unauthorized execution is induced by a preceding untrusted Observation, it constitutes a successful indirect prompt injection attack within the scope of this paper.
We therefore focus on whether a candidate call is authorized when it crosses the real execution boundary, rather than on whether the Agent reads, responds to, or uses external content.

This problem cannot be solved by preventing all Observation-derived information from participating in subsequent execution, because legitimate actions and arguments in open-ended tool tasks can often be instantiated only using runtime information
\cite{li2026agentdyn}.
The key is therefore not simply to reduce trust in Observations, but to distinguish whether runtime information is \emph{instantiating existing authority (runtime instantiation)} or \emph{expanding existing authority (authority expansion)}.
To this end, a runtime authorization mechanism must maintain the following three properties during task execution.

\smallskip \noindent\textbf{Authority Non-Escalation.}
Observations and information produced during runtime execution may supply the objects, identifiers, and environmental facts required to complete an existing task, but they cannot independently add operations, targets, recipients, permission scopes, or external effects not authorized by the user.
Accordingly, even when subsequent execution depends on an Observation, real tool calls must still satisfy the principles of least privilege and complete mediation~\cite{1451869}.

\smallskip \noindent\textbf{Persistent Action Origin.}
If an action or argument is initially introduced through operational semantics in an untrusted Observation, its provenance attribute must not disappear automatically merely because of subsequent normal execution, contextual propagation, or the passage of time.
This requirement is consistent with the principle in classical information-flow control and dynamic taint tracking that provenance attributes remain persistent: a security attribute should not be implicitly cleared merely because it passes through normal processing steps
\cite{DBLP:journals/cacm/Denning76,DBLP:journals/jsac/SabelfeldM03,10.1145/2619091}.

\smallskip \noindent\textbf{Origin Non-Override.}
Subsequent legitimate execution may add new facts and execution evidence for a dynamic object, but an existing action origin cannot be overridden merely because the same value later reappears in the execution history.
Historical reappearance may add evidence, but it cannot independently create authority that did not previously exist; for an argument already associated with an action-inducing origin, historical recurrence alone cannot override that origin or promote the argument into execution authority.

Accordingly, this paper studies the following problem:
\textbf{How can a system allow untrusted Observations to remain involved in solving open-ended tasks while maintaining the above authorization properties at the real tool-execution boundary, such that runtime information can instantiate existing user authority but cannot thereby acquire execution authority that the user never granted?}

\subsection{Threat Model}
\label{sec:system-threat-model}

\smallskip \noindent\textbf{System and Deployment Model.}
SARA is deployed between the Agent and the real tool executor; Observations may enter the Agent's context normally and influence planning, but real candidate calls must pass runtime control before execution.
SARA neither replaces the Agent's task planning nor depends on hidden chains of thought or other unexposed internal states.
SARA scopes its authorization state to a single task execution associated with one user request and clears the runtime authorization state established for that task when the task terminates or is explicitly reset.

\smallskip \noindent\textbf{Trust Assumptions.}
We trust the integrity of the user request \(U\), the tool schema \(\mathcal{T}\), the SARA runtime, and the real tool executor; the user request serves as the task's authorization root, while the tool schema defines the accessible interfaces and argument structure.
Webpages, emails, documents, search results, API responses, and all other tool Observations are treated as untrusted input.
Trust in SARA means only that the attacker cannot directly modify its code, model configuration, runtime state, or execution logic; it does not assume that SARA's semantic judgments are necessarily correct, and an unauthorized real call that SARA erroneously allows because of an attacker-controlled Observation still counts as a defense failure.

\smallskip \noindent\textbf{Attacker Capabilities and Goal.}
An attacker may control part of the natural-language content or structured fields in one or more tool Observations and embed explicit operational requests, disguised instructions, tool-use suggestions, conditional steps, or attacker-specified argument values, following the standard setting in indirect prompt injection research
\cite{greshake2023indirect,zhan2024injecagent,debenedetti2024agentdojo}.
This content may influence the Agent's subsequent planning and form a real candidate call after several normal queries, reads, or other tool steps.
The attacker's goal is to cause a call induced by external content to produce a real external effect not authorized by the user; however, the attacker cannot modify the user request, trusted tool schema, SARA runtime, or real tool executor, nor can the attacker bypass SARA to submit a tool call directly.

\smallskip \noindent\textbf{Protection Goal and Scope.}
We protect tool behaviors with real external effects, including sending or disclosing information, creating or modifying external state, changing access permissions, and committing transactions or reservations.
An attack is counted as successful only if a call induced by an untrusted Observation ultimately crosses the runtime control and produces a real external effect not authorized by the user; merely reading or discussing attack content, generating a malicious plan, producing a candidate call that is ultimately rejected, or performing reads, queries, or object-resolution operations that do not themselves cause unauthorized external effects does not independently count as attack success.
We allow normal tasks to depend on facts and dynamic objects in Observations, so mere data dependence is outside the protection target.
If a user explicitly asks the Agent to unconditionally execute arbitrary subsequent instructions in external content, thereby actively delegating decision authority to that content itself, the case is outside our scope; if external content determines only local data or runtime objects while the operation type, target, or external effect remains constrained by the user request, the case remains within scope.

\section{Related Work}

\smallskip \noindent\textbf{Indirect Prompt Injection and Dynamic Tool Execution.}
Indirect prompt injection (IPI) allows attackers to influence the subsequent behavior of LLM applications through operational content embedded in webpages, emails, or other external data\cite{greshake2023indirect}.
Prior work has further systematized and benchmarked prompt-injection attack forms and defenses \cite{299563}.
InjecAgent extended this threat to Agent settings with real tool semantics \cite{zhan2024injecagent}, while AgentDojo further jointly evaluates user tasks and attack goals through executable workflows, extending the security consequences of IPI from textual responses to real tool execution \cite{debenedetti2024agentdojo}; broader evaluations also show that the risk spans multiple runtime stages, including tool use, execution, and external feedback \cite{ye-etal-2024-toolsword,zhang2025asb}.
The recently proposed AgentDyn\cite{li2026agentdyn} further highlights the security--utility tension in dynamic environments: legitimate tasks may themselves depend on runtime feedback and third-party information to determine subsequent behavior; tool-use benchmarks such as $\tau$-bench and ToolSandbox also include dynamic interaction, stateful execution, and inter-tool dependencies, requiring tasks to be further instantiated using state obtained during execution \cite{yao2024taubenchbenchmarktoolagentuserinteraction,lu-etal-2025-toolsandbox}.
Dynamic tool tasks therefore require defenses that neither simply block the influence of Observations on subsequent execution nor permit attack semantics within them to propagate without constraint.

\smallskip \noindent\textbf{From Input Control to Action Attribution.}
Existing IPI defenses have gradually evolved from limiting the influence of external content on Agents to constraining executable behavior and analyzing the origins from which concrete tool actions are formed.
Early input- and model-side defenses primarily strengthen trust boundaries: StruQ separates prompts and data into structured channels, Instruction Hierarchy reinforces high-privilege instructions through instruction priorities, and SecAlign improves model robustness against injected instructions through preference optimization \cite{307694,wallace2024instructionhierarchytrainingllms,chen2025secalign}.
PI Detector, Spotlighting, and Prompt Sandwiching address untrusted input through injection detection, provenance marking, and trusted-instruction reinforcement, respectively \cite{protectai2024deberta,hines2024spotlighting,schulhoff2024sandwich}.
As tool-using Agents have become more capable, Tool Filter, IPIGuard, Task Shield, and CaMeL have further constrained execution deviation through tool-action filtering, execution-structure constraints, task consistency, and separation of trusted control flow from untrusted data flow, respectively \cite{debenedetti2024agentdojo,an2025ipiguard,jia2025taskshield,debenedetti2025camel}.
Recent work has further examined the origins of candidate actions: MELON compares tool behavior across contexts through masked re-execution, whereas AttriGuard uses action-level causal attribution and counterfactual execution to analyze the causal contributions of user intent and untrusted Observations to candidate actions \cite{zhu2025melon,he2026attriguard}.
However, in dynamic tool tasks, runtime data such as file identifiers, object IDs, and third-party responses may be obtainable only through interaction and may legitimately drive subsequent tool calls; an action being ``driven by an Observation'' therefore identifies only its formation origin and does not directly determine whether it has execution authority.

\paragraph{Runtime Authorization and Provenance Control.}
Recent work has begun to directly examine whether a candidate call has sufficient execution authority at the real tool-execution boundary.
RTBAS adapts Information Flow Control to tool-based Agents and constrains integrity and confidentiality in tool calls through runtime dependency analysis, preventing untrusted information flows from directly driving sensitive operations \cite{DBLP:journals/corr/abs-2502-08966}.
ClawGuard derives task-level rules from user goals and enforces runtime constraints before tool calls to prevent candidate behaviors outside the authorization scope from producing real external effects \cite{zhao2026clawguard}.
PACT refines authorization to the argument level, assigns distinct authority semantics to tool arguments, and propagates runtime-value provenance across execution steps, allowing authorization decisions for concrete argument bindings to retain their original provenance after intermediate data flows \cite{fan2026pact}.
AuthGraph pre-generates an Authorization Graph from the user request and tool set in an isolated clean context, encoding expected tool sequences and allowed source tools for security-critical arguments, and detects deviations through structural alignment with actual execution provenance; for tasks that depend on runtime information, its authorization graph can be extended in a restricted manner at prespecified replanning points, while new paths remain constrained by the pre-authorized tool set \cite{wang2026authgraph}.
AIRGuard distinguishes runtime information from real execution authority through action-time authority control and determines whether a current side effect remains within the permitted scope by jointly considering task-to-step authority, source/target trust, and cross-step risk information \cite{qin2026airguard}.

These works advance the security of tool-using Agents from input filtering and behavioral analysis toward runtime authorization at the execution boundary.
Building on them, we focus on a finer-grained provenance distinction: an external Observation may both provide action-inducing provenance that forms a candidate action and contain dynamic information obtained during legitimate task execution, but the latter must form execution evidence through a trusted execution process before it can support subsequent authorization.
The key is therefore not merely to track runtime provenance, but to simultaneously preserve two distinct forms of evidence---``what external content induced'' and ``what legitimate execution established''---and prevent action-inducing provenance from being implicitly promoted to new execution authority after cross-step propagation or historical reappearance.

\begin{figure*}[t]
    \centering
    \includegraphics[width=\linewidth]{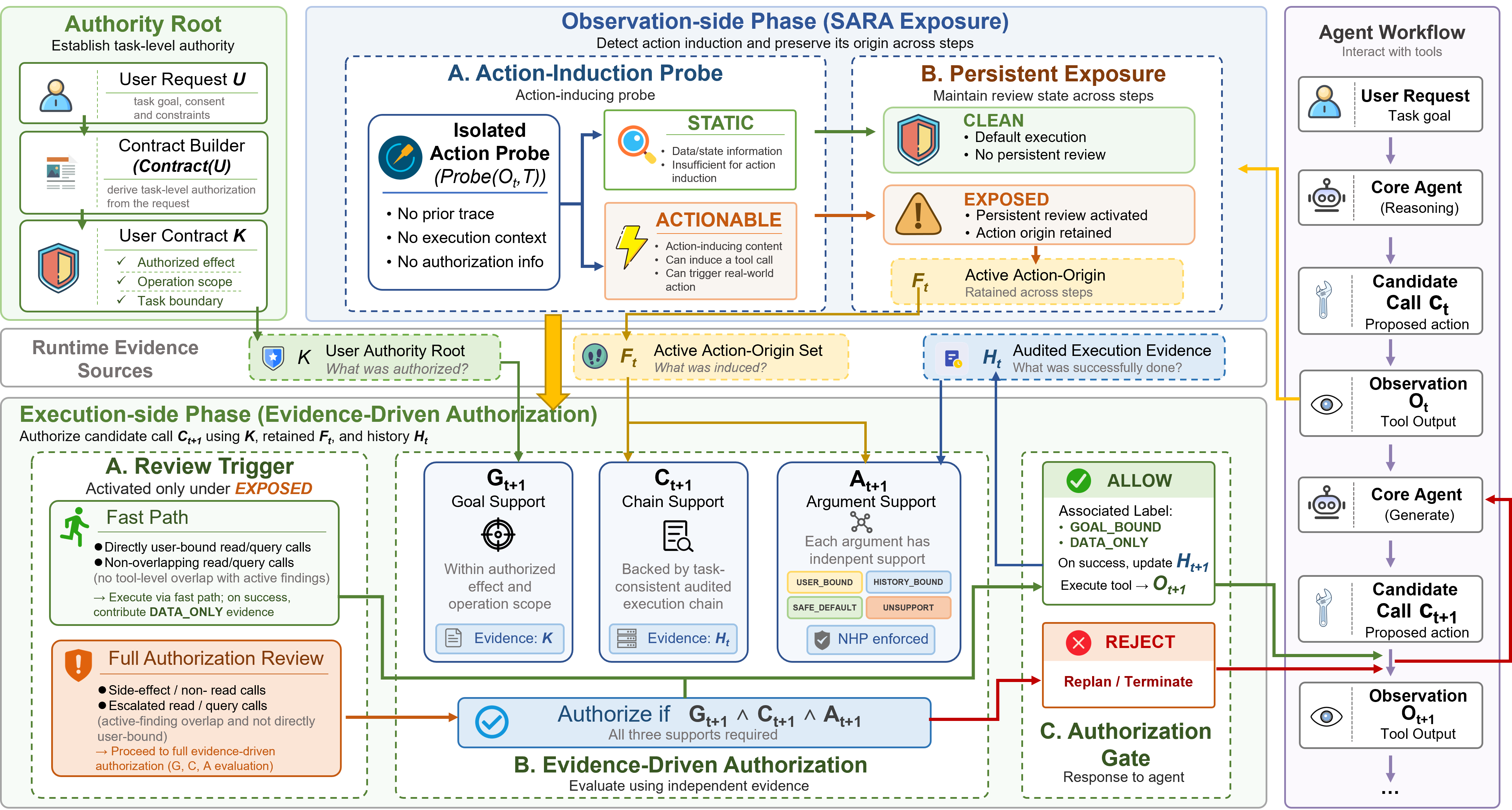}
    \caption{SARA's runtime authorization framework. The user request establishes the authorization root \(K\); the Observation-side Action Probe identifies action-inducing semantics and persistently maintains action origins \(F_t\), which trigger persistent review but do not grant execution authority; authorized and successful tool executions accumulate audited execution evidence \(H_t\). At the real tool-execution boundary, SARA jointly uses \(K\), \(F_t\), and \(H_t\) to authorize the goal, execution chain, and arguments of a candidate call.}
    \label{fig:sara-overview}
\end{figure*}

\section{SARA: Separating Action Induction from Runtime Authorization}
\label{sec:method}

\subsection{Design Overview}
\label{sec:sara-overview}

Completing open-ended tasks typically requires tool-using Agents to use runtime information from Observations, such as file identifiers and order numbers; SARA therefore allows Observations to continue participating in task instantiation while separating their influence on candidate-action formation from real execution authorization.
Its core principle is
\[
    \text{Action Induction} \not\Rightarrow \text{Execution Authorization}.
\]

SARA organizes runtime control into two distinct but connected stages around this principle.
At the Observation-side stage, a context-isolated Action Probe does not determine whether external content is malicious; it only detects whether the content expresses action-inducing semantics and, when possible, maps them to concrete tool actions or argument roles.
Ordinary facts or state information leave the trajectory \texttt{CLEAN}; once an Observation is classified as \texttt{ACTIONABLE}, the trajectory transitions to \texttt{EXPOSED}, the corresponding action and argument origins are retained across steps as active action origins, and persistent authorization review is activated.
At the execution-side stage, candidate calls with real side effects under \texttt{EXPOSED} undergo full authorization review; query and read calls are escalated for review only when they are related to existing Probe findings and cannot be directly determined from user authorization, while all other calls continue along the fast path.
Full review independently determines whether a candidate call has sufficient execution support based on user authorization, persistent action origins, and audited execution evidence; the Probe therefore records only action induction and the resulting review obligation, without granting or denying real execution authority.

To support this control, SARA maintains three complementary runtime information sources: the user authorization root \(K\) represents the task authority boundary established by the trusted user request, the persistent action-origin set \(F_t\) records tool actions or argument roles previously induced by external Observations, and the audited execution history \(H_t\) records allowed and successfully completed tool interactions to provide positive execution evidence for legitimate dynamic dependencies.
These sources correspond to the runtime authorization properties defined in Section~\ref{sec:problem}: \(K\) provides an independent authority root for \emph{Authority Non-Escalation}, \(F_t\) maintains \emph{Persistent Action Origin}, and \(F_t\), \(H_t\), together with the subsequent No-History-Promotion mechanism, maintain \emph{Origin Non-Override} to prevent execution history from overriding an existing action origin.
Thus, \(F_t\) and \(H_t\) preserve distinct evidentiary semantics established by action induction and legitimate execution, respectively, and neither can expand the user authorization boundary defined by \(K\).
Figure~\ref{fig:sara-overview} summarizes this runtime control flow.

\subsection{Constructing the Authorization Root}
\label{sec:sara-contract}

SARA first constructs a task-level authorization contract from the user's original request \(U\):
\[
    K=\mathrm{Contract}(U),
\]
where each contract item is represented as
\[
    k_j=
    \left(
    e_j,\,
    \mathrm{operation}_j,\,
    \mathrm{scope}_j,\,
    \Sigma_j
    \right),
\]
where \(e_j\) denotes an effect type permitted by the user, \(\mathrm{operation}_j\) denotes an allowed operation, \(\mathrm{scope}_j\) denotes the objects, recipients, or external scope on which the operation may act, and \(\Sigma_j\) denotes static arguments explicitly supplied by the user.
The contract \(K\) defines the task-level upper bound on authority without attempting to enumerate the complete tool-call sequence in advance.
This design preserves the runtime flexibility required by semi-open or open-ended tasks: users can typically specify a semantic goal and the external effects they permit, but cannot provide in advance all object identifiers, environmental states, or unique execution paths needed to complete the task.
Accordingly, \(K\) always remains the independent \emph{authority root} for subsequent execution decisions; runtime information may only instantiate existing authority and cannot create new task authority.

\subsection{Action Induction and Persistent Provenance Retention}
\label{sec:sara-exposure}

\smallskip \noindent\textbf{Context-Isolated Action Probe.}
After establishing an independent authorization root, SARA identifies on the Observation side whether external content exhibits \emph{action-inducing semantics} capable of concretely affecting subsequent tool behavior.
This stage neither determines whether the external content is malicious nor whether an action it proposes is authorized by the user; it asks only: \emph{Can this Observation itself express an action or argument role that maps to the current tool interfaces?}
Its core criterion is that an Observation has action-inducing semantics if, when isolated from the constraints of the user goal, it remains sufficient to independently prompt the model to instantiate a specific tool behavior.
For each tool output \(O_t\), SARA executes under context isolation
\[
    (z_t,Q_t,M_t)
    =
    \mathrm{Probe}(O_t,\mathcal{T}),
\]
where \(\mathcal{T}\) is the tool set in the current environment, \(z_t\in\{\texttt{STATIC},\texttt{ACTIONABLE}\}\) indicates whether the Observation has action-inducing semantics, \(Q_t\) denotes the tool-level action footprint instantiated from it, and \(M_t\) denotes the set of argument-origin anchors instantiated as tool arguments.
If the Observation explicitly urges an action, then $z_t=\texttt{ACTIONABLE}$; when the action can be mapped to the current tool set, the corresponding $Q_t$ and $M_t$ are generated. Explicit but unmappable or incompletely specified actions may still trigger \texttt{ACTIONABLE} without producing mapped action-origin anchors. Otherwise, $z_t=\texttt{STATIC}$.

Each argument-origin anchor \(m\in M_t\) directly instantiated by an Observation has the form

\[
    m = (\tau,p,\mathrm{norm}(v)),
\]

where \(\tau\) denotes the tool, \(p\) denotes the argument path, and \(\mathrm{norm}(v)\) denotes the normalized argument value.
While \(Q_t\) describes the tool behavior proposed by external content, \(M_t\) further records the argument roles and concrete values that participated in that behavior, distinguishing ``a value appeared in an Observation'' from ``a value was proposed as part of an external action.''
For example, an email address that appears only as contact information is ordinary runtime data, whereas the same address in an operational request to ``forward the report to this address'' creates an action origin bound to the send action and recipient argument.
The Probe therefore records \emph{action-origin provenance} bound to concrete tool actions and argument roles, rather than general data lineage as described in conventional provenance tracking~\cite{soton356851}.
\texttt{ACTIONABLE} is likewise not a label of maliciousness or authorization: legitimate operational instructions from third parties may also have action-inducing semantics, so SARA uses the label only as a signal to trigger subsequent persistent authorization review, not as grounds to directly reject an Observation or candidate call.

\smallskip \noindent\textbf{Persistent Review and Action-Origin Retention.}
Action induction formed in an Observation does not necessarily appear immediately as the next real tool call; the Agent may perform several searches, reads, or object-resolution steps before reusing an action, target, or argument proposed in an earlier Observation.
Checking only the local transition \(O_t\rightarrow c_{t+1}\) therefore cannot maintain action-origin constraints, and an established action origin must persist across subsequent tool steps.
To this end, SARA introduces a trajectory state \(S_t\in\{\texttt{CLEAN},\texttt{EXPOSED}\}\) to model the persistent review obligation:
\[
    S_t=
    \begin{cases}
        \texttt{EXPOSED}, & \exists\, i\in\{1,\dots,t\},\,z_i=\texttt{ACTIONABLE}, \\
        \texttt{CLEAN},   & \text{otherwise}.
    \end{cases}
\]

\texttt{EXPOSED} does not mean that an attack has occurred or presuppose that subsequent calls are malicious; it indicates only that external content with action-inducing semantics has appeared in the trajectory and that subsequent execution must therefore maintain a persistent review obligation.
In sync with this state, SARA maintains an Active Action-Origin Set \(F_t\).
Let the current action origin be \(\phi_t=(Q_t,M_t)\), and let \(F_0=\varnothing\); then
\[
    F_t=
    \begin{cases}
        F_{t-1}\cup\{\phi_t\}, & z_t=\texttt{ACTIONABLE}, \\
        F_{t-1},               & z_t=\texttt{STATIC}.
    \end{cases}
\]

Once an action origin is written to \(F_t\), ordinary Observations, normal intermediate calls, or a subsequent reappearance of the same value in the execution history do not automatically clear it, allowing the origin to persist across multiple tool steps in \(O_i\rightarrow\cdots\rightarrow c_{t+1}\).
This design is consistent with the idea in classical information-flow control that security flow constraints persist through normal computation~\cite{10.1145/1323293.1294293}, but \(F_t\) stores action origins bound to concrete tool actions and argument roles rather than general information-flow labels.
This persistence does not constitute a permanent denylist: \(F_t\) records only the actions or argument roles through which external content participated in forming behavior and does not directly decide whether a subsequent candidate call may execute.

\subsection{Evidence-Driven Authorization at the Execution Boundary}
\label{sec:sara-authorization}

When the main Agent proposes a real candidate call \(c_{t+1}\) after observing \(O_t\), SARA shifts control from Observation-side action induction to execution-side authorization and evaluates the current call using the user authorization root \(K\), persistent action origins \(F_t\), and audited execution evidence \(H_t\).
Runtime control can be summarized as
\[
    (K,F_t,H_t,c_{t+1})
    \longrightarrow
    \mathrm{Auth}_{t+1},
\]
where \(F_t\) records what external content previously induced and \(H_t\) records what has been established through calls that were allowed and actually executed; neither can replace \(K\) to create new user authority.

\smallskip \noindent\textbf{Review Trigger.}
SARA treats \texttt{ACTIONABLE} as a trigger for persistent review, not as an attack verdict.
While the task trajectory remains \texttt{CLEAN}, the system has not observed action-inducing external content, so full authorization review is not enabled and candidate calls follow the default execution path.
Once the trajectory enters \texttt{EXPOSED}, SARA applies differentiated review to subsequent calls: all candidate calls with real side effects enter the full authorization path, whereas query and read calls are escalated only when their tool identity has tool-level overlap with a Probe action footprint retained in \(F_t\) and their actual arguments cannot be directly determined from the user request or authorization contract.
All remaining calls continue along the fast path, and their successful results participate in subsequent authorization decisions only as \texttt{DATA\_ONLY} evidence.
Thus, \texttt{EXPOSED} represents a persistent and differentiated review obligation, rather than an attack state or global block.

\smallskip \noindent\textbf{Audited Execution Evidence.}
Dynamic objects in open-ended tasks are often obtainable only through runtime interaction, so full authorization must use positive evidence established during execution; however, an Observation cannot thereby become a new authority source.
SARA records only tool interactions that were previously allowed and actually executed successfully as positive audited evidence.
Let the initial state be \(H_0=\varnothing\); the audited execution history up to time \(t\) is defined as
\[
    H_t=
    \left\{
    (c_i,O_i,\ell_i)
    \mid
    1\le i\le t,\;
    \mathrm{Allow}(c_i)
    \land
    \mathrm{Success}(c_i)
    \right\},
\]
where \(\ell_i\) describes the evidentiary capacity of the successful execution and its result for subsequent authorization and is assigned according to its runtime control path as
\[
    \ell_i=
    \begin{cases}
        \texttt{DATA\_ONLY},  & \text{\texttt{EXPOSED} fast path}, \\
        \texttt{GOAL\_BOUND}, & \text{otherwise}.
    \end{cases}
\]

\texttt{DATA\_ONLY} permits an execution result to propagate as runtime data and instantiate arguments within the existing authorization scope, but it cannot independently establish a new recipient, permission, amount, or other external effect.
\texttt{GOAL\_BOUND} retains execution relations associated with the current authorized task and may further participate in execution-chain and argument-binding decisions.
Rejected candidate calls, failed calls, plans expressed in the Agent's natural language, and ordinary restatements of contextual content cannot acquire audited-execution-evidence status merely by entering the history.
For a runtime value \(v\), a legitimate dynamic binding can be represented as
\[
    U
    \overset{K}{\Longrightarrow}
    c_i
    \xrightarrow{\mathrm{exec}}
    O_i
    \ni v
    \xrightarrow{\mathrm{bind}}
    a_{t+1},
    \qquad
    a_{t+1}\in\theta_{t+1}.
\]
This relation requires \(v\) to originate from a real successful execution serving the current authorized task and then to instantiate an argument role in the existing task; SARA therefore does not accept a value merely because it appeared in the context or history; instead, it requires an auditable, task-consistent execution origin.
In this way, \(H_t\) allows dynamic objects that the user cannot enumerate in advance to participate in task instantiation while remaining bounded by the authority boundary defined by \(K\).

\smallskip \noindent\textbf{Goal, Chain, and Argument Support.}
For a candidate call \(c_{t+1}\) entering the full authorization path, SARA does not require the complete tool sequence or argument-provenance path to be enumerated in advance; instead, it incrementally establishes three complementary forms of execution support online, based on the current authorization contract \(K\), persistent action-origin set \(F_t\), and audited execution history \(H_t\).
\[
    \begin{aligned}
        G_{t+1}(c) & = \mathrm{GoalSupport}(c,U,K),      \\
        C_{t+1}(c) & = \mathrm{ChainSupport}(c,H_t),     \\
        A_{t+1}(c) & = \mathrm{ArgSupport}(c,U,H_t,F_t).
    \end{aligned}
\]

\(G_{t+1}(c)\) checks whether the current call's operation, target, and scope remain within the user-task boundary defined by \(K\), preventing runtime information from expanding external effects that the user did not authorize.
\(C_{t+1}(c)\) checks whether the runtime objects or state on which the current call depends can be established through task-consistent successful execution relations in \(H_t\), rather than merely because a value appeared in ordinary context.
\(A_{t+1}(c)\) checks the binding basis of every actual argument in its current tool-semantic role.

For each argument, SARA classifies its support source as \texttt{USER\_BOUND}, \texttt{HISTORY\_BOUND}, \texttt{SAFE\_DEFAULT}, or \texttt{UNSUPPORTED} according to user authorization, audited execution history, active action origins, and tool semantics.
\texttt{USER\_BOUND} means that the argument can be directly determined from the user request or authorization contract; \texttt{HISTORY\_BOUND} means that it can be established by a task-consistent successful execution chain; \texttt{SAFE\_DEFAULT} means that it is a default binding that does not expand the task's authority scope; and \texttt{UNSUPPORTED} means that the argument currently lacks acceptable independent support.
Both \(C_{t+1}\) and \(A_{t+1}\) may use \(H_t\), but the former constrains the execution chain on which the candidate call as a whole depends, whereas the latter constrains the binding provenance of a concrete argument value in its corresponding semantic role.

\smallskip \noindent\textbf{No-History-Promotion.}
The same argument value may first be recorded in \(F_t\) as an action argument in an \texttt{ACTIONABLE} Observation and later reappear in \(H_t\) through normal tool interactions; the later occurrence may add factual evidence but cannot, merely through historical reappearance, erase the existing action-inducing origin.
To maintain \emph{Origin Non-Override}, SARA applies No-History-Promotion (NHP) when determining argument support.
If a current argument matches an argument-origin anchor preserved in \(F_t\), a subsequent reappearance of the value in \(H_t\) cannot independently grant it \texttt{HISTORY\_BOUND} support or allow it to bypass the existing provenance constraint through \texttt{SAFE\_DEFAULT}.
If the argument has independent support from the user request or authorization contract, it may still be classified as \texttt{USER\_BOUND}; otherwise, it remains \texttt{UNSUPPORTED}.
NHP therefore prevents the implicit provenance-promotion process
\[
    \text{action origin}
    \rightarrow
    \text{historical reappearance}
    \rightarrow
    \text{authority}
\]
rather than permanently prohibiting Observation-derived values.

\smallskip \noindent\textbf{Authorization Decision and Evidence Commit.}
For a candidate call entering the full authorization path, SARA requires all three support conditions to hold:
\[
    \mathrm{Auth}_{t+1}(c_{t+1})
    =
    G_{t+1}(c_{t+1})
    \land
    C_{t+1}(c_{t+1})
    \land
    A_{t+1}(c_{t+1}).
\]
Here, \(A_{t+1}\) incorporates NHP's constraint on provenance conflicts, so a candidate call can pass the full authorization path only if it simultaneously satisfies the user-goal boundary, a task-consistent execution chain, and concrete argument-binding requirements.
When \(\mathrm{Auth}_{t+1}(c_{t+1})=1\), SARA returns \texttt{ALLOW} and submits the candidate call to the real tool executor; if sufficient support cannot be established, it returns \texttt{REJECT} before any real external effect occurs and feeds the rejection result with its reason back to the main Agent for replanning.
Only tool interactions that are allowed and actually execute successfully can commit new positive evidence to subsequent state.
If \(c_{t+1}\) executes successfully and returns \(O_{t+1}\), then
\[
    H_{t+1}
    =
    H_t
    \cup
    \left\{
    (c_{t+1},O_{t+1},\ell_{t+1})
    \right\},
\]
where \(\ell_{t+1}\) is determined by the control path that the call actually followed; if the call is rejected or execution fails, no new audited execution evidence is produced.

Subsequently, \(O_{t+1}\) is processed as the next Observation by the Action Probe, which updates \(S_{t+1}\) and \(F_{t+1}\) accordingly.
SARA thereby forms a runtime control loop:
\[
    O_t
    \!\rightarrow\!
    (S_t,F_t)
    \!\rightarrow\!
    c_{t+1}
    \!\rightarrow\!
    \mathrm{Auth}_{t+1}
    \!\rightarrow\!
    \mathrm{Exec}
    \!\rightarrow\!
    (O_{t+1},H_{t+1}).
\]

This loop maintains two non-interchangeable runtime states: \(F_t\) preserves action origins established by external Observations, whereas \(H_t\) preserves positive evidence established by previously authorized and successfully executed interactions; neither can expand the user authorization root \(K\).
SARA therefore neither takes over the Agent's task planning nor prevents Observations from participating in normal reasoning; under persistent review, it prevents action-inducing external content that lacks independent authorization support from being converted into real external effects.

\section{Experimental Setup}
\label{sec:evaluation-environments}

We evaluate SARA by systematically addressing the following four research questions:
\begin{itemize}
    \item \textbf{RQ1: Can SARA suppress unauthorized tool execution while preserving task utility?} (~\ref{sec:rq1_main})
    \item \textbf{RQ2: Do SARA's security gains remain consistent across different Agent backbones?} (~\ref{sec:rq2_cross_backbone})
    \item \textbf{RQ3: How do SARA's key runtime mechanisms contribute to security and task utility?} (~\ref{sec:rq3_ablation})
    \item \textbf{RQ4: What additional inference cost does SARA incur to achieve its security--utility gains?} (~\ref{sec:rq4_cost})
\end{itemize}

\smallskip \noindent\textbf{Benchmarks and attack protocol.}
We evaluate SARA on AgentDojo~\cite{debenedetti2024agentdojo} and AgentDyn~\cite{li2026agentdyn}, which cover standard IPI workflows and dynamic tool tasks that depend more heavily on runtime feedback, respectively.
AgentDojo contains four executable suites---Banking, Slack, Travel, and Workspace---whereas AgentDyn contains Shopping, GitHub, and DailyLife and further introduces dynamic planning, cross-application interaction, and third-party environmental feedback.
We follow each benchmark's protocol for combining user and injection tasks and use four IPI formulations while keeping the attack goal fixed: \texttt{ignore\_previous} (IP)~\cite{perez2022ignorepreviouspromptattack} overrides existing instructions, \texttt{system\_message} (SM)~\cite{debenedetti2024agentdojo} disguises the attack as a system instruction, \texttt{important\_instructions} (II)~\cite{debenedetti2024agentdojo} reinforces a malicious request through an importance claim, and \texttt{tool\_knowledge} (TK)~\cite{debenedetti2024agentdojo} presents it as tool knowledge or operational guidance.
The four formulations share the same attack goals to reduce the dependence of results on a single attack template.
AgentDojo contains 92 benign tasks and 3,528 attack instances, with 882 instances for each attack formulation; AgentDyn contains 141 benign tasks and 5,202 attack instances, with 1,338 instances each for IP, SM, and II and 1,188 for TK.
All methods use the same combinations of user tasks, attack tasks, and attack formulations to ensure instance-level comparability.

\smallskip \noindent\textbf{Metrics.}
We evaluate task utility without attacks, task utility under attacks, and attack success rate to separate security from the ability to complete user tasks.
Let \(\mathcal{B}\) and \(\mathcal{A}\) denote the sets of benign and attack instances, respectively, with \(N_{\mathrm{B}}=|\mathcal{B}|\) and \(N_{\mathrm{A}}=|\mathcal{A}|\).
For instance \(i\), let \(\mathrm{User}_i\) indicate that the user-task success condition ultimately holds and let \(\mathrm{Attack}_i\) indicate that the attack goal is ultimately judged successful through real environment state or the corresponding benchmark grader; then
\begin{gather*}
    \mathrm{BU} = \frac{\sum_{i\in\mathcal{B}} \mathbb{I}\left[\mathrm{User}_i\right]}{N_{\mathrm{B}}}, \qquad
    \mathrm{UA} = \frac{\sum_{i\in\mathcal{A}} \mathbb{I}\left[\mathrm{User}_i\right]}{N_{\mathrm{A}}} \\[6pt]
    \mathrm{ASR} = \frac{\sum_{i\in\mathcal{A}} \mathbb{I}\left[\mathrm{Attack}_i\right]}{N_{\mathrm{A}}}
\end{gather*}

\textbf{Benign Utility (BU)} measures the user-task completion rate without attacks, \textbf{Utility under Attack (UA)} measures the user-task completion rate in the presence of attacks, and \textbf{Attack Success Rate (ASR)} measures the fraction of attack goals that ultimately succeed.
Instances with runtime errors remain in the denominator of the evaluation set, but rejected or invalid candidate calls and calls that do not produce the real target effect are not counted as attack success; if a malicious effect has already taken effect in the real environment, subsequent execution errors do not revoke that success.
All real tool calls are processed according to the corresponding benchmark's scoring criteria, and task or attack state changes only when a call successfully reaches the tool executor and produces an actual environmental effect.
All comparison methods use identical user requests, tool sets, environment states, scoring criteria, and attack instances, with method-specific defense mechanisms configured according to their official implementations.
UA and ASR are weighted over attack instances; because the number of benign tasks is relatively small, BU is reported as the mean over three independent runs to reduce small-sample variation.

\smallskip \noindent\textbf{Baselines.}
We use the undefended Agent as the base reference and select methods that cover different security-control positions and provide representative comparisons for the problem addressed by SARA.
PI Detector~\cite{debenedetti2024agentdojo} and Spotlighting~\cite{hines2024spotlighting} represent input-side detection and provenance-marking methods, allowing us to test whether identifying or attenuating untrusted instructions in advance is sufficient to mitigate IPI without changing authorization at the execution stage.
Tool Filter~\cite{debenedetti2024agentdojo}, IPIGuard~\cite{an2025ipiguard}, and CaMeL~\cite{debenedetti2025camel} represent structured execution-constraint methods that constrain the execution space through restricting available tools, preplanning tool dependencies, and separating trusted control flow from untrusted data flow, respectively, enabling comparison between preemptively narrowing execution structure and SARA's runtime authorization.
MELON~\cite{zhu2025melon} and AttriGuard~\cite{he2026attriguard} determine whether a candidate action is primarily driven by an untrusted Observation through re-execution or counterfactual analysis, and thus provide comparisons with SARA's action-origin tracking and independent authorization decision.
Finally, ClawGuard~\cite{zhao2026clawguard} and AIRGuard~\cite{qin2026airguard} both enforce runtime control before real tool calls and are the direct comparison methods closest to SARA's control position.
This collection therefore covers input filtering, execution-structure constraints, action attribution, and execution-boundary control, with particular emphasis on representative mechanisms that test SARA's core design choices.

\subsection{RQ1: End-to-End Security and Task Utility}
\label{sec:rq1_main}

In RQ1, we evaluate all comparison methods using GPT-4o-mini~\cite{openai2024gpt4omini} and Gemini-2.5-Flash-Lite~\cite{google2025gemini25flashlite} as the two main Agent backbones to examine end-to-end security and task utility across different foundation models.
Table~\ref{tab:main_results} shows that SARA substantially reduces ASR in all four primary evaluation settings across AgentDojo and AgentDyn while maintaining UA no lower than the corresponding Agent-only baseline.
With GPT-4o-mini, ASR decreases from \(15.79\%\) and \(16.07\%\) to \(0.06\%\) and \(0.17\%\) on AgentDojo and AgentDyn, respectively; with Gemini-2.5-Flash-Lite, it decreases from \(33.28\%\) and \(30.91\%\) to \(0.62\%\) and \(0.63\%\), respectively.
Because UA and ASR capture performance only under attack, we further use BU to measure the structural utility cost that a defense imposes on legitimate runtime dependencies in attack-free sessions.

\begin{table*}[htbp]
    \centering
    \caption{End-to-end performance comparison across models on AgentDojo and AgentDyn. BU and UA are higher-is-better ($\uparrow$), while ASR is lower-is-better ($\downarrow$).}
    \label{tab:main_results}
    \renewcommand{\arraystretch}{1.0}

    \begin{tabular*}{\textwidth}{@{\extracolsep{\fill}}lcccccc|cccccc}
        \toprule
        \multicolumn{1}{c}{\multirow{3}{*}{\textbf{Defense}}} & \multicolumn{6}{c}{\textbf{GPT-4o-mini}} & \multicolumn{6}{c}{\textbf{Gemini-2.5-Flash-Lite}} \\
        \cmidrule(lr){2-7} \cmidrule(lr){8-13}
        & \multicolumn{3}{c}{\textbf{AgentDojo}} & \multicolumn{3}{c}{\textbf{AgentDyn}} & \multicolumn{3}{c}{\textbf{AgentDojo}} & \multicolumn{3}{c}{\textbf{AgentDyn}} \\
        \cmidrule(lr){2-4} \cmidrule(lr){5-7} \cmidrule(lr){8-10} \cmidrule(lr){11-13}
        & \textbf{BU} $\uparrow$ & \textbf{UA} $\uparrow$ & \textbf{ASR} $\downarrow$ & \textbf{BU} $\uparrow$ & \textbf{UA} $\uparrow$ & \textbf{ASR} $\downarrow$ & \textbf{BU} $\uparrow$ & \textbf{UA} $\uparrow$ & \textbf{ASR} $\downarrow$ & \textbf{BU} $\uparrow$ & \textbf{UA} $\uparrow$ & \textbf{ASR} $\downarrow$ \\
        \midrule
        Agent-only & \underline{72.46} & 56.01 & 15.79 & \textbf{65.01} & 54.58 & 16.07 & \textbf{87.68} & 57.43 & 33.28 & \textbf{78.96} & 58.77 & 30.91 \\
        PI Detector & 44.20 & 25.57 & 3.94 & 30.02 & 19.49 & 3.83 & 47.10 & 25.71 & 5.84 & 30.50 & 19.42 & 5.17 \\
        Spotlighting & \textbf{73.19} & 59.84 & 13.29 & \underline{64.78} & \underline{54.88} & 13.61 & \underline{86.23} & 65.08 & 26.16 & \underline{75.18} & \underline{64.92} & 22.97 \\
        Tool Filter & 66.67 & \underline{63.27} & 2.01 & 44.21 & 45.29 & 1.83 & 39.49 & 47.25 & 1.62 & 24.35 & 28.47 & 1.10 \\
        IPIGuard & 69.93 & 62.36 & 1.39 & 42.55 & 43.62 & 0.71 & 72.46 & 61.31 & 2.58 & 36.88 & 35.70 & 1.19 \\
        CaMeL & 40.94 & 46.28 & \underline{0.11} & 25.30 & 31.16 & \textbf{0.12} & 71.74 & \textbf{77.89} & \textbf{0.28} & 49.41 & 53.27 & \textbf{0.21} \\
        MELON & 63.77 & 48.61 & 0.85 & 48.94 & 40.68 & 0.69 & 80.43 & 48.33 & 0.71 & 58.63 & 36.54 & \underline{0.48} \\
        AttriGuard & 59.06 & 52.92 & 0.48 & 47.04 & 46.16 & 1.02 & 80.43 & 76.98 & 0.68 & 64.30 & \textbf{66.26} & 1.17 \\
        ClawGuard & 68.84 & 60.66 & 4.54 & 52.24 & 47.90 & 4.29 & 84.06 & 76.59 & 7.00 & 65.48 & 63.90 & 6.56 \\
        AIRGuard & 65.58 & 49.72 & 13.32 & 54.85 & 45.06 & 13.57 & 79.35 & 52.61 & 27.01 & 67.85 & 50.54 & 25.62 \\
        \midrule
        SARA (Ours) & 66.67 & \textbf{63.44} & \textbf{0.06} & 56.03 & \textbf{54.92} & \underline{0.17} & 81.16 & \underline{77.27} & \underline{0.62} & 67.14 & \underline{64.92} & 0.63 \\
        \bottomrule
    \end{tabular*}
\end{table*}

\smallskip \noindent\textbf{Security--Utility Trade-offs with Existing Defenses.}
Different defense mechanisms exhibit markedly different security--utility trade-offs.
Input-side methods generally interfere less with normal execution but may leave a high residual ASR; for example, Spotlighting maintains a UA of \(64.92\%\) on AgentDyn with Gemini-2.5-Flash-Lite, but its ASR remains \(22.97\%\).
Tool Filter and IPIGuard achieve lower ASR by narrowing the tool or execution space earlier, but incur more pronounced utility costs on dynamic tasks: on AgentDyn with GPT-4o-mini, their ASRs are \(1.83\%\) and \(0.71\%\), respectively, while their UAs are only \(45.29\%\) and \(43.62\%\).
CaMeL uses stronger execution isolation, yielding ASRs of \(0.11\%\)--\(0.28\%\) across the four settings and lower ASR than SARA in three settings, but it also incurs a more substantial benign utility cost.
Relative to Agent-only, CaMeL reduces BU by \(31.52\), \(39.71\), \(15.94\), and \(29.55\) percentage points across the four settings, whereas SARA's reductions are only \(5.79\), \(8.98\), \(6.52\), and \(11.82\) percentage points.
This difference indicates that strong execution isolation can further reduce ASR but may also continuously restrict normal runtime dependencies; SARA instead allows raw Observations to participate in task execution and concentrates security control on authorizing real candidate calls after action induction activates review.

Action-attribution methods also exhibit different boundaries.
On AgentDyn with Gemini-2.5-Flash-Lite, AttriGuard achieves a UA of \(66.26\%\), slightly higher than SARA's \(64.92\%\), but its ASR of \(1.17\%\) remains higher than SARA's \(0.63\%\), indicating that an Observation's contribution to forming an action does not directly determine the action's ultimate execution authority.
Similarly, ClawGuard leaves a residual ASR of \(4.29\%\)--\(7.00\%\) across the four settings, whereas AIRGuard yields an ASR of \(13.32\%\)--\(27.01\%\) together with lower UA in every setting.
Overall, these results indicate that constraining the execution space, identifying the origin of action formation, or imposing higher-level task and trust constraints alone may still struggle to preserve legitimate runtime dependencies while blocking specific unauthorized execution; SARA further authorizes calls at the execution boundary by jointly considering action origins, audited execution evidence, and actual argument bindings.

SARA is not uniformly best on every individual metric; for example, CaMeL obtains lower ASR in three settings.
However, across all four primary evaluation settings, SARA holds ASR at or below \(0.63\%\), maintains UA no lower than Agent-only, and achieves these security gains with a substantially smaller benign utility cost.
RQ1 therefore indicates that SARA's advantage does not come from unconditionally compressing runtime behavior to pursue the lowest ASR, but from achieving a more consistent end-to-end security--utility balance while preserving Observation-dependent task execution.

\subsection{RQ2: Security Gains and Utility across Backbones}
\label{sec:rq2_cross_backbone}

\begin{figure}[htbp]
    \centering
    \begin{subfigure}{0.49\linewidth}
        \centering
        \includegraphics[width=\linewidth]{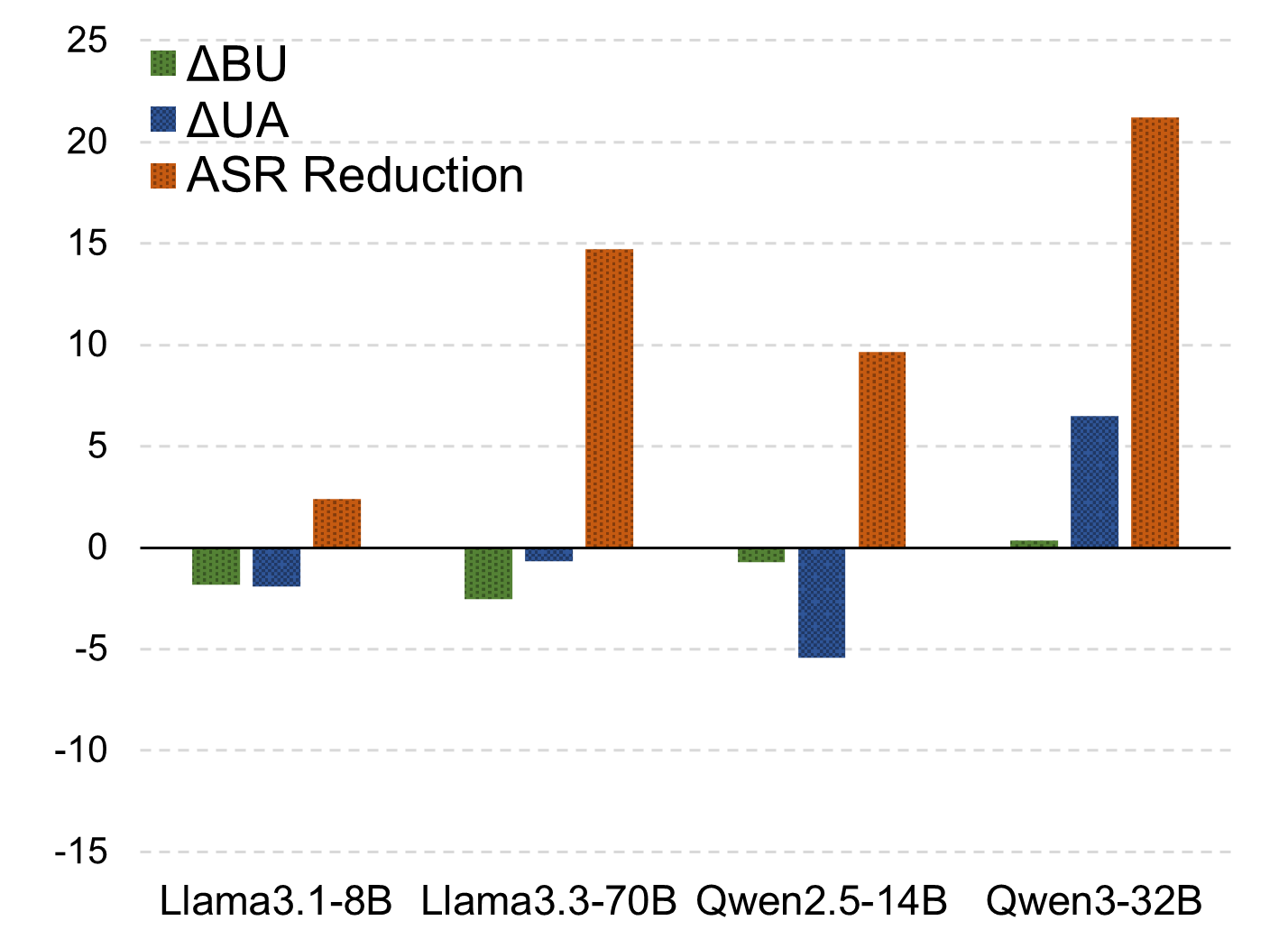}
        \caption{AgentDojo}
        \label{fig:models-dojo}
    \end{subfigure}
    \hfill
    \begin{subfigure}{0.49\linewidth}
        \centering
        \includegraphics[width=\linewidth]{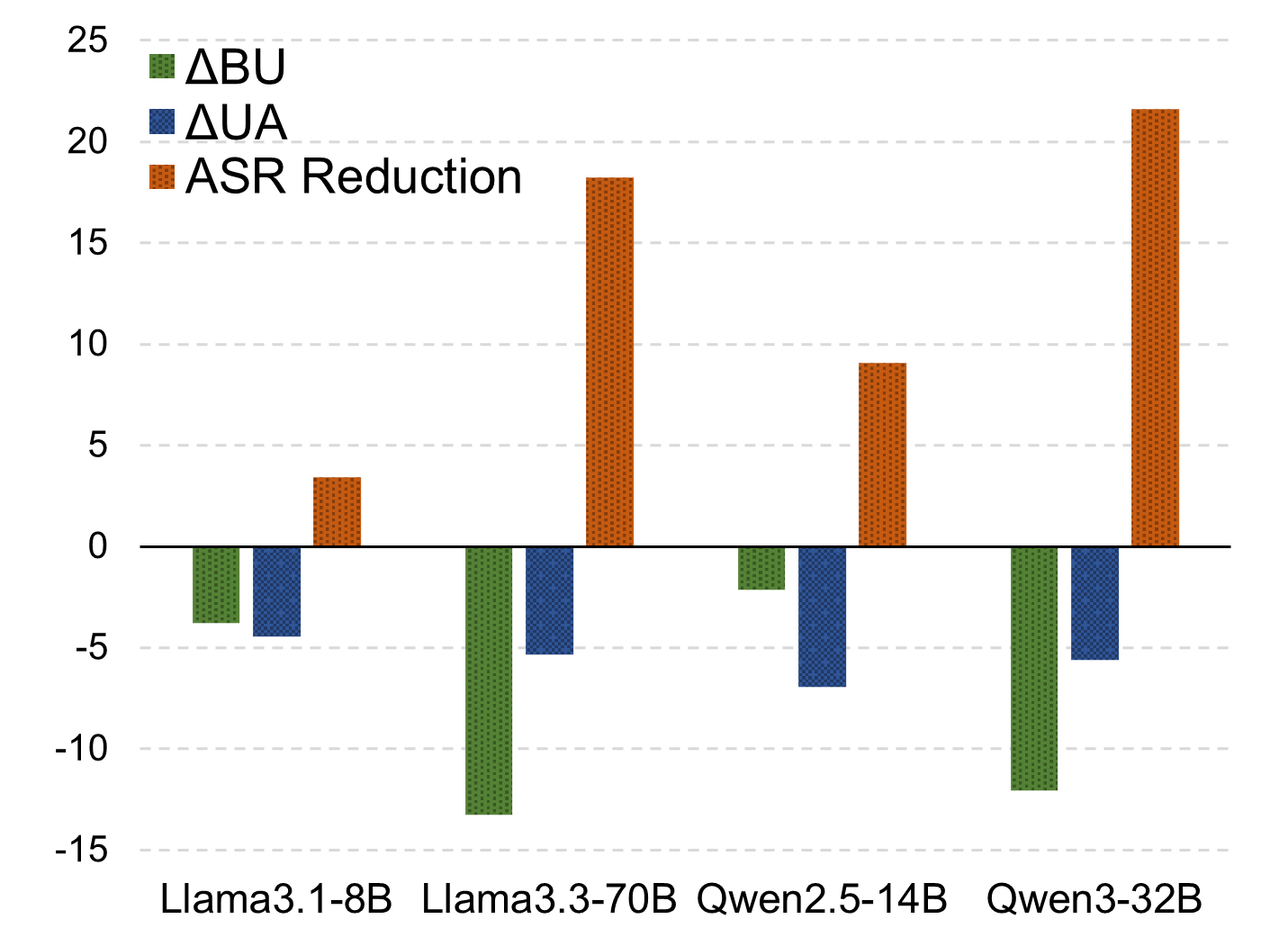}
        \caption{AgentDyn}
        \label{fig:models-dyn}
    \end{subfigure}

    \caption{Cross-backbone changes in BU, UA, and ASR from Agent-only to SARA on AgentDojo and AgentDyn.}
    \label{fig:models}
\end{figure}

RQ2 selects Llama3.1-8B-Instruct~\cite{llama2024herd}, Llama3.3-70B-Instruct~\cite{meta2024llama33}, Qwen2.5-14B~\cite{qwen2_5_tech_report}, and Qwen3-32B~\cite{qwen3_tech_report}, covering different model families, parameter scales, and native tool-use capabilities, and compares Agent-only with SARA on AgentDojo and AgentDyn to examine SARA's performance across open-weight backbones of varying scales; SARA's runtime mechanisms and evaluation protocol remain unchanged across all backbones.
Table~\ref{tab:rq2_absolute} summarizes the results, and Figure~\ref{fig:models} presents SARA's performance under each backbone.

Across all $4 \times 2 = 8$ backbone--benchmark combinations, ASR decreases substantially after integrating SARA, indicating that its security gains are not limited to a single main Agent in our evaluation.
On AgentDojo, Llama3.1-8B's ASR decreases from $4.02\%$ to $1.64\%$, while the other three backbones all fall to at most $0.03\%$; on AgentDyn, Llama3.1-8B leaves a residual ASR of $1.75\%$, while the other three backbones all remain below $0.3\%$.
Thus, although the backbones differ markedly in their attack exposure under Agent-only, SARA consistently narrows the success space for unauthorized execution, and its cross-backbone consistency is reflected primarily in security rather than absolute task utility.
In contrast to the consistent reduction in ASR, BU and UA are more sensitive to the backbone and task environment.
On AgentDojo, BU changes by no more than $2.53$ percentage points for all four backbones, while UA changes range from a decrease of $5.42$ to an increase of $6.49$ percentage points, indicating that task utility after security intervention depends more strongly on the main Agent's execution, replanning, and recovery capabilities in a given workflow.
On AgentDyn, which depends more heavily on dynamic objects, third-party information, and runtime feedback, both BU and UA decrease for all four backbones; UA decreases by $4.44$--$6.94$ percentage points, while BU decreases by $13.24$ and $12.06$ percentage points for Llama3.3-70B and Qwen3-32B, respectively, suggesting that complex dynamic workflows amplify the cost of maintaining task continuity after security intervention.

\begin{table}[htbp]
    \centering
    \caption{Cross-backbone BU, UA, and ASR on AgentDojo and AgentDyn. Each cell reports \emph{Agent-only $\rightarrow$ SARA} (\%).}
    \label{tab:rq2_absolute}
    \setlength{\tabcolsep}{0pt}
    \renewcommand{\arraystretch}{1.1}
    \begin{tabular*}{\columnwidth}{@{\extracolsep{\fill}}l ccc@{}}
        \toprule
        \multicolumn{1}{c}{\textbf{Backbone}} & \textbf{BU} $\uparrow$ & \textbf{UA} $\uparrow$ & \textbf{ASR} $\downarrow$ \\
        \midrule
        \rowcolor{gray!15}
        \multicolumn{4}{c}{\textbf{AgentDojo}} \\
        Llama3.1-8B    & 34.42$\rightarrow$32.61 & 25.96$\rightarrow$24.04 & 4.02$\rightarrow$1.64 \\
        Llama3.3-70B   & 55.07$\rightarrow$52.54 & 43.91$\rightarrow$43.23 & 14.71$\rightarrow$0.00 \\
        Qwen2.5-14B    & 52.90$\rightarrow$52.17 & 49.55$\rightarrow$44.13 & 9.64$\rightarrow$0.00 \\
        Qwen3-32B      & 63.04$\rightarrow$63.41 & 46.54$\rightarrow$53.03 & 21.26$\rightarrow$0.03 \\
        \midrule
        \rowcolor{gray!15}
        \multicolumn{4}{c}{\textbf{AgentDyn}} \\
        Llama3.1-8B    & 25.77$\rightarrow$21.99 & 20.13$\rightarrow$15.69 & 5.19$\rightarrow$1.75 \\
        Llama3.3-70B   & 51.54$\rightarrow$38.30 & 40.79$\rightarrow$35.43 & 18.32$\rightarrow$0.10 \\
        Qwen2.5-14B    & 41.37$\rightarrow$39.24 & 39.81$\rightarrow$32.87 & 9.30$\rightarrow$0.25 \\
        Qwen3-32B      & 56.03$\rightarrow$43.97 & 46.64$\rightarrow$41.04 & 21.74$\rightarrow$0.13 \\
        \bottomrule
    \end{tabular*}
\end{table}

Execution logs further explain this utility divergence.
Llama3.1-8B has a \texttt{tool\_error} rate of 19\%--35\%, substantially higher than the 7\%--12\% of the Qwen models and the 11\%--17\% of Llama3.3-70B, primarily due to failures in generating structured tool calls or execution errors; this observation is consistent with prior Agent benchmarks showing that weaker models fail more often in multi-step interaction and complex tool use~\cite{liu2025agentbenchevaluatingllmsagents,NEURIPS2024_877b4068}.
This also explains why Llama3.1-8B simultaneously exhibits low UA and anomalously low Agent-only ASR on both benchmarks: many task trajectories terminate because of execution failures before an attack can produce a real effect, so lower ASR does not indicate stronger defensive capability.
SARA can block candidate calls that lack authorization, but it cannot replace the main Agent's subsequent planning and error recovery; when the backbone lacks the ability to recover after interception, security intervention is more likely to manifest as lower BU or UA.

Overall, RQ2 shows that SARA's cross-backbone consistency is primarily reflected in security: its runtime authorization mechanism continuously reduces ASR across main Agents and task environments, whereas task utility depends more strongly on the backbone's own execution and recovery capabilities and the dynamism of the workflow.
Together with the GPT-4o-mini and Gemini-2.5-Flash-Lite results in RQ1, these findings suggest that Agents with stronger execution and recovery capabilities are more likely to maintain or even improve utility under attack while keeping ASR low.
SARA therefore provides a runtime authorization boundary that is relatively consistent across backbones, rather than a mechanism for improving task utility independently of the main Agent's capabilities.

\subsection{RQ3: Contributions of Key Runtime Mechanisms}
\label{sec:rq3_ablation}

This section answers RQ3 through end-to-end ablation experiments that evaluate how SARA's key runtime states and authorization constraints contribute to security and task utility.
All variants keep user tasks, attack instances, tool environments, and other runtime logic identical; however, because authorization decisions can alter the Agent's subsequent planning path, the results measure each mechanism's end-to-end effect on the complete execution process.



\begin{table}[htbp]
    \centering
    \caption{End-to-end component ablation results for SARA on AgentDojo and AgentDyn. AIT: Action-Induction Tracking; AE: Audited Evidence; PSG: Parameter-Support Gate; NHP: No-History-Promotion.}
    \label{tab:ablation}
    \setlength{\tabcolsep}{0pt}
    \renewcommand{\arraystretch}{1.05}

    \begin{tabular*}{\columnwidth}{@{\extracolsep{\fill}}l ccc ccc@{}}
        \toprule
        \multicolumn{1}{c}{\raisebox{-1.25ex}[0pt][0pt]{\textbf{Variant}}} & \multicolumn{3}{c}{\textbf{AgentDojo}} & \multicolumn{3}{c}{\textbf{AgentDyn}} \\
        \cmidrule(lr){2-4}\cmidrule(lr){5-7}
        & \textbf{BU} $\uparrow$ & \textbf{UA} $\uparrow$ & \textbf{ASR} $\downarrow$ & \textbf{BU} $\uparrow$ & \textbf{UA} $\uparrow$ & \textbf{ASR} $\downarrow$ \\
        \midrule
        \textbf{SARA} & 66.67 & 63.44 & 0.06 & 56.03 & 54.92 & 0.17 \\
        w/o AIT & 67.39 & 62.05 & 3.37 & 56.50 & 53.21 & 3.02 \\
        w/o AE & 68.12 & 53.49 & 0.00 & 49.17 & 42.10 & 0.12 \\
        w/o PSG & 69.20 & 64.54 & 2.49 & 59.10 & 56.50 & 2.73 \\
        w/o NHP & 67.75 & 64.80 & 0.65 & 56.26 & 54.63 & 0.75 \\
        \bottomrule
    \end{tabular*}
\end{table}

As shown in Table~\ref{tab:ablation}, we construct four ablation variants:
\textit{w/o Action-Induction Tracking} (AIT): disables the Observation-side Action Probe and cross-step action-origin tracking, so subsequent authorization no longer has access to provenance evidence in \(F_t\), while side-effecting calls are still reviewed against the user authorization contract and execution history and read calls are allowed only as \texttt{DATA\_ONLY} evidence.
\textit{w/o Audited Evidence} (AE): removes the positive audited execution evidence provided by \(H_t\).
\textit{w/o Parameter-Support Gate} (PSG): continues to compute argument-level support \(A_t(c)\) but removes its final veto, reducing \(G_t(c)\land C_t(c)\land A_t(c)\) to \(G_t(c)\land C_t(c)\).
\textit{w/o No-History-Promotion} (NHP): removes only the No-History-Promotion constraint applied when an action origin conflicts with subsequent historical evidence.

\smallskip
\noindent\textbf{AIT (Cross-Step Induction Tracking).}
Without AIT, \(F_t\) is no longer maintained, and ASR rises from \(0.06\%\) and \(0.17\%\) to \(3.37\%\) and \(3.02\%\) on AgentDojo and AgentDyn, respectively, while BU changes by less than \(1\) percentage point on both benchmarks.
This result shows that AIT primarily contributes by persistently retaining the action-inducing origin of earlier Observations, preventing subsequent candidate calls from acquiring apparent legitimacy solely from the current task context or execution history.
Without this persistent action-origin state, the authorizer can still use \(K\) and \(H_t\), but cannot determine whether a current call continues an earlier external action-inducing source, thereby substantially weakening its ability to identify cross-step attacks.

\smallskip
\noindent\textbf{PSG (Argument-Level Veto).}
Removing the final veto imposed by \(A_t(c)\) increases both BU and UA on the two benchmarks, but also raises ASR to \(2.49\%\) and \(2.73\%\), revealing a clear security--utility trade-off.
This result shows that satisfying \(G_t(c)\) and \(C_t(c)\) alone is insufficient to guarantee that a call is authorized, because an attack call can preserve the overall task direction and execution chain while introducing localized authority expansion only through concrete argument bindings, such as recipients or target objects.
PSG therefore enforces the authorization boundary at the actual argument level, at the cost that some legitimate calls with ambiguous authorization boundaries may also be rejected more conservatively.

\smallskip
\noindent\textbf{AE (Positive Audited Evidence).}
Removing \(H_t\) does not increase ASR: it decreases from \(0.06\%\) and \(0.17\%\) to \(0.00\%\) and \(0.12\%\) on AgentDojo and AgentDyn, respectively.
However, UA drops by approximately \(10\) and \(13\) percentage points, and BU on AgentDyn decreases from \(56.03\%\) to \(49.17\%\).
This result shows that the primary role of AE is not to further tighten the security boundary, but to provide an auditable legitimate execution source for runtime-generated dynamic values, such as file identifiers and object IDs.
Without \(H_t\), these dynamic bindings---which do not come directly from user input but are necessary for progressing open-ended tasks---have difficulty obtaining \texttt{HISTORY\_BOUND} support, making authorization substantially more conservative.
Thus, \(F_t\) preserves external action-inducing origins, whereas \(H_t\) provides positive support for legitimate runtime dependencies; together, they prevent persistent review from degenerating into indiscriminate rejection of dynamic execution.

\smallskip
\noindent\textbf{NHP (Protection against Provenance Laundering).}
Removing NHP raises ASR to \(0.65\%\) and \(0.75\%\) on AgentDojo and AgentDyn, respectively, while BU and UA show no consistent change, consistent with its targeted role in handling specific provenance conflicts.
NHP applies only when an argument value first appears as an action origin and later reappears in \(H_t\), preventing this historical reappearance alone from promoting the value to \texttt{HISTORY\_BOUND} support.
Because such conflicts account for only a subset of runtime bindings, NHP has a much smaller impact on utility than PSG, while its security contribution lies in blocking the provenance-laundering path ``action origin \(\rightarrow\) historical reappearance \(\rightarrow\) authority promotion.''

Overall, RQ3 shows that SARA's security--utility balance arises from the asymmetric roles of these four mechanisms.
AIT, PSG, and NHP primarily maintain the security boundary over action origins, argument bindings, and provenance conflicts, whereas AE establishes positive runtime evidence through audited successful executions, preventing these constraints from becoming overly restrictive.
SARA's complete authorization mechanism therefore does not rely on a single stricter criterion; instead, it jointly preserves ``what external content induced'' and ``what legitimate execution has established,'' and resolves potential provenance conflicts between them at the argument level.

\subsection{RQ4: Security--Utility Gains and Additional Inference Cost}
\label{sec:rq4_cost}

\begin{figure}[htbp]
    \centering
    \begin{subfigure}{0.49\linewidth}
        \centering
        \includegraphics[width=\linewidth,trim=5 12 5 5, clip]{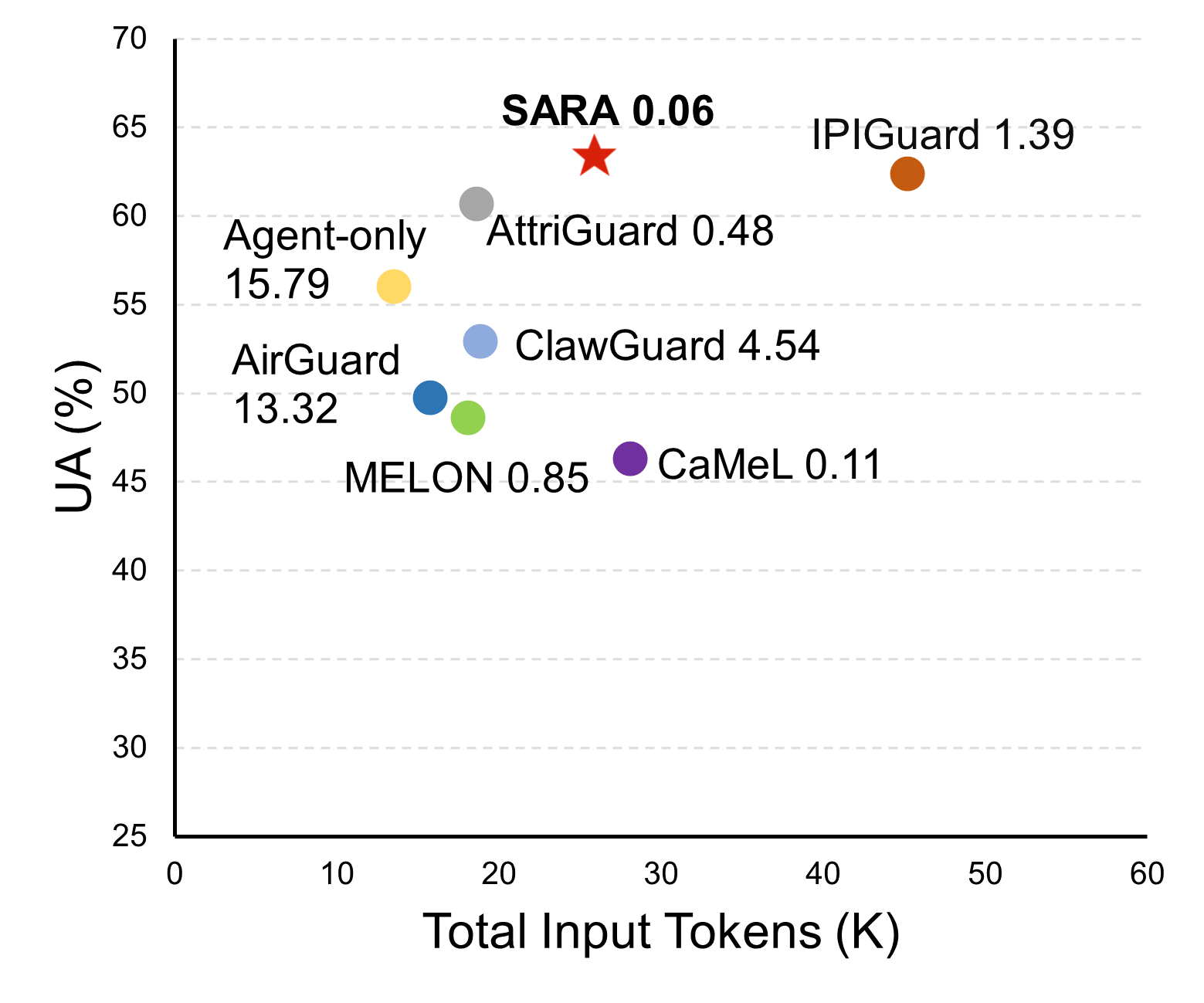}
        \caption{AgentDojo}
        \label{fig:tradeoff-dojo}
    \end{subfigure}
    \hfill
    \begin{subfigure}{0.49\linewidth}
        \centering
        \includegraphics[width=\linewidth,trim=0 12 5 5, clip]{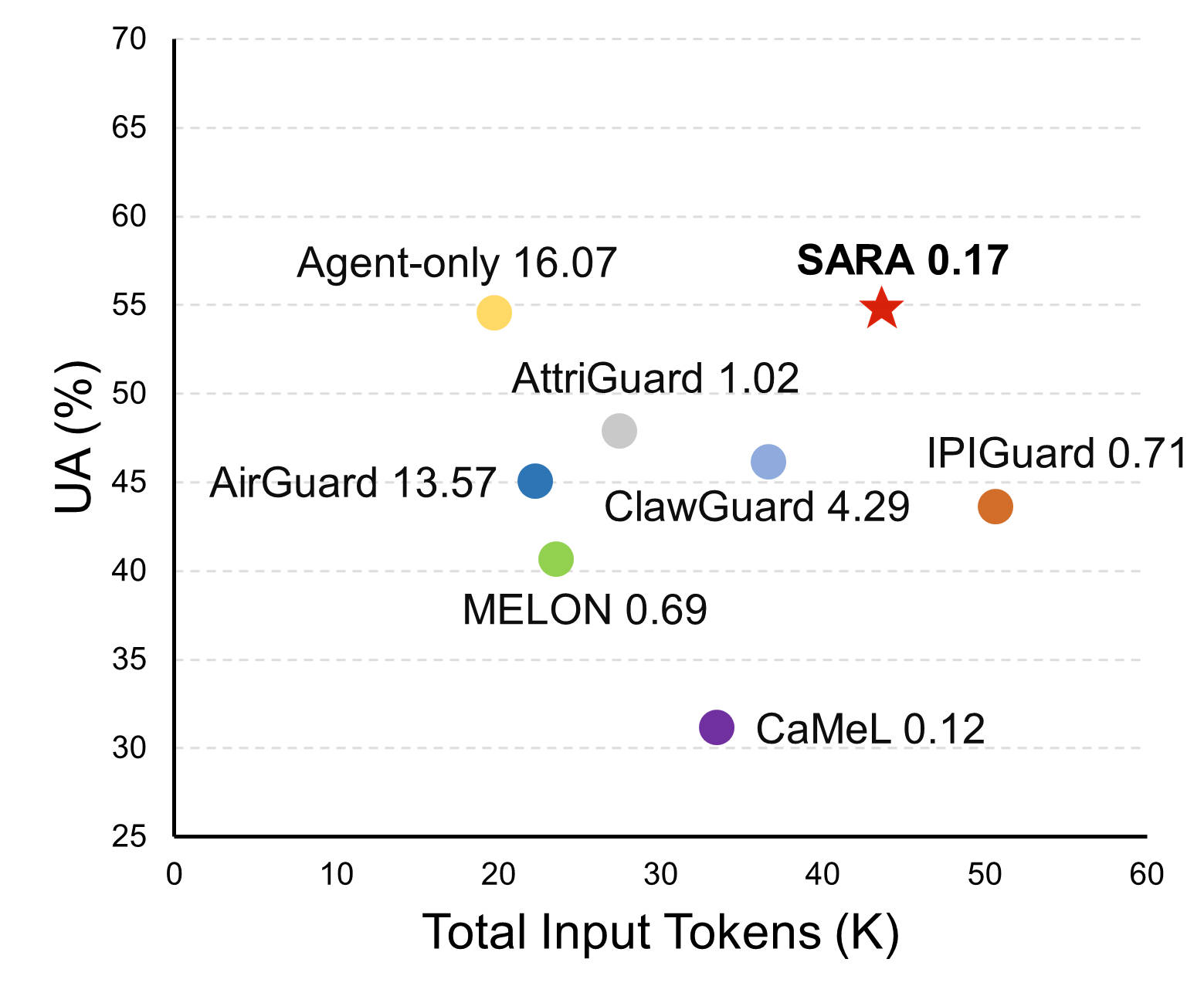}
        \caption{AgentDyn}
        \label{fig:tradeoff-dyn}
    \end{subfigure}

    \caption{Security--utility--cost trade-off on AgentDojo and AgentDyn. Numbers beside methods denote ASR (\%).}
    \label{fig:tradeoff}
\end{figure}

RQ4 further evaluates the runtime inference cost that SARA incurs to obtain the above security and utility gains, as well as the composition of that cost.
Because end-to-end latency is readily affected by network conditions, model scheduling, and deployment configurations, we use average token consumption and API-call counts on GPT-4o-mini attack tasks as the primary cost metrics.
Table~\ref{tab:token-usage-gpt4omini} decomposes the cost between the Agent and Guard, while Figure~\ref{fig:tradeoff} combines total input cost, UA, and ASR to show the security--utility--cost operating points of different defenses.

\begin{table}[t]
    \centering
    \caption{Runtime inference cost on attack tasks across AgentDojo and AgentDyn.}
    \label{tab:token-usage-gpt4omini}
    \setlength{\tabcolsep}{0pt} 
    \begin{tabular*}{\columnwidth}{@{\extracolsep{\fill}}lcccccc}
        \toprule
        \multicolumn{1}{c}{\raisebox{-1.35ex}[0pt][0pt]{\textbf{Method}}} & \multicolumn{2}{c}{\textbf{Agent}} & \multicolumn{2}{c}{\textbf{Guard}} & \multicolumn{2}{c}{\textbf{Total}} \\
        \cmidrule(lr){2-3}\cmidrule(lr){4-5}\cmidrule(lr){6-7}
        & \textbf{Input} & \textbf{Calls} & \textbf{Input} & \textbf{Calls} & \textbf{Input} & \textbf{Output} \\
        \midrule
        \rowcolor{gray!15}
        \multicolumn{7}{c}{\textbf{AgentDojo}} \\
        Agent-only & 13.54K & 4.25 & -- & -- & 13.54K & 0.28K \\
        IPIGuard & 36.91K & 7.51 & 8.31K & 1.99 & 45.22K & 0.97K \\
        CaMeL & 27.70K & 4.57 & 0.39K & 0.80 & 28.10K & 1.50K \\
        MELON & 9.62K & 3.41 & 8.47K & 5.13 & 18.09K & 0.42K \\
        AttriGuard & 10.57K & 4.21 & 8.32K & 6.76 & 18.89K & 1.38K \\
        ClawGuard & 15.73K & 5.07 & 2.93K & 2.45 & 18.65K & 0.30K \\
        AIRGuard & 15.00K & 4.38 & 0.77K & 2.98 & 15.77K & 0.48K \\
        \textbf{SARA} & 17.01K & 4.87 & 8.85K & 2.42 & 25.86K & 0.93K \\
        \midrule
        \rowcolor{gray!15}
        \multicolumn{7}{c}{\textbf{AgentDyn}} \\
        Agent-only & 19.74K & 5.55 & -- & -- & 19.74K & 0.28K \\
        IPIGuard & 40.76K & 8.08 & 9.90K & 1.99 & 50.66K & 1.03K \\
        CaMeL & 33.04K & 5.14 & 0.39K & 0.85 & 33.43K & 1.63K \\
        MELON & 12.41K & 4.01 & 11.14K & 7.87 & 23.54K & 0.45K \\
        AttriGuard & 20.83K & 6.25 & 15.79K & 9.76 & 36.62K & 1.82K \\
        ClawGuard & 23.90K & 6.98 & 3.55K & 3.19 & 27.45K & 0.32K \\
        AIRGuard & 21.51K & 5.72 & 0.73K & 2.95 & 22.24K & 0.47K \\
        \textbf{SARA} & 25.75K & 6.55 & 17.86K & 4.91 & 43.61K & 1.28K \\
        \bottomrule
    \end{tabular*}
\end{table}

SARA is not the defense with the lowest token overhead, but its additional computational investment corresponds to a more competitive security--utility operating point.
On AgentDojo, SARA's average Agent and Guard inputs are 17.01K and 8.85K, respectively, for a total input of 25.86K, or \(1.91\times\) that of Agent-only, corresponding to a UA of \(63.44\%\) and an ASR of \(0.06\%\).
Some methods have lower input costs but are generally accompanied by higher residual ASR or lower task utility.
On the more dynamic AgentDyn, SARA's total input increases to 43.61K, or \(2.21\times\) that of Agent-only, corresponding to a UA of \(54.92\%\) and an ASR of \(0.17\%\), only slightly higher than CaMeL's \(0.12\%\) ASR.
Thus, when the three objectives are lower total input cost, higher UA, and lower ASR, SARA forms a competitive non-dominated operating point on both benchmarks rather than obtaining security gains by pursuing the lowest token overhead.

SARA's additional overhead arises primarily from two stages.
The first is the \textbf{direct cost of authorization analysis}: SARA invokes its Guard components for authorization-root construction, tool-effect classification, Observation probing, and execution-time authorization review, corresponding to Guard Input/Calls in the table.
The second is the \textbf{indirect cost of trajectory adaptation}: SARA is responsible only for execution authorization and does not replace the main Agent's planning; a rejected candidate call returns control to the Agent for replanning and therefore incurs additional Agent-side inference.
Relative to Agent-only, SARA increases Agent Input/Calls from \(13.54\text{K}/4.25\) to \(17.01\text{K}/4.87\) on AgentDojo and from \(19.74\text{K}/5.55\) to \(25.75\text{K}/6.55\) on AgentDyn.
Both Guard- and Agent-side costs are higher on AgentDyn, consistent with its greater dependence on runtime information: more dynamic objects and environmental feedback increase both the execution context required for authorization decisions and the planning burden of finding a legitimate path after rejection.

Overall, RQ4 shows that SARA's security gains incur additional inference cost.
On both benchmarks, SARA trades additional computation for very low ASR and high UA, indicating that its advantage lies in the system-level trade-off among security, task utility, and runtime cost rather than in minimizing computational overhead.

\section{Conclusion and Limitations}
\label{sec:conclusion_limitations}

This work introduces SARA, a runtime authorization mechanism designed to mitigate unauthorized tool execution caused by untrusted Observations while preserving support for legitimate runtime adaptation.
SARA decouples \emph{action induction} from \emph{execution authorization}.
Specifically, it persistently tracks the provenance of Observation-induced actions, re-evaluates authorization at the tool boundary based on user authorization and audited execution evidence, and prevents historical reappearance alone from promoting action-origin provenance into execution authority.
Evaluations on AgentDojo and AgentDyn demonstrate that SARA limits the Attack Success Rate (ASR) to no more than $0.63\%$ across the four primary evaluation settings, delivering consistent security gains across diverse Agent backbones.

These results come with two main limitations.
First, SARA functions as a runtime authorization mechanism rather than a formal security guarantee.
Its semantic judgments may yield false positives or false negatives, and recovering from blocked executions relies on the host Agent's replanning capabilities, making overall utility dependent on model capability and workflow dynamics.
Second, our empirical evaluation is restricted to tool-based Indirect Prompt Injection (IPI) workflows under the threat model defined in Section~\ref{sec:system-threat-model}, which assumes trusted user inputs, tool schemas, the SARA runtime, and the underlying tool executor.
Consequently, SARA does not address direct attacks bypassing the authorization runtime, pure data-dependency vulnerabilities, or unconditional control delegation to external sources.

\bibliographystyle{plainurl}
\bibliography{references}

\end{document}